\documentclass[letterpaper]{article} 
\usepackage[submission]{aaai2026}  
\usepackage{times}  
\usepackage{helvet}  
\usepackage{courier}  
\usepackage[hyphens]{url}  
\usepackage{graphicx} 
\usepackage{natbib}  
\usepackage{caption} 
\usepackage{algorithm}
\usepackage{algorithmic}
\usepackage{tabularx}
\usepackage{minted}
\usepackage{microtype}
\usepackage{multirow}
\usepackage{graphicx}
\usepackage{subfigure}
\usepackage{booktabs} 
\usepackage[utf8]{inputenc} 
\usepackage[T1]{fontenc}    
\usepackage{csquotes}
\usepackage{subfig}
\usepackage{amsmath}
\usepackage{amssymb}
\usepackage{mathtools}
\usepackage{amsthm}
\usepackage{url}            
\usepackage{booktabs}       
\usepackage{amsfonts}       
\usepackage{nicefrac}       
\usepackage{microtype}      
\usepackage{xcolor}         
\usepackage{subfiles}

\usepackage{newfloat}
\usepackage{listings}
\DeclareCaptionStyle{ruled}{labelfont=normalfont,labelsep=colon,strut=off} 
\floatstyle{ruled}
\newfloat{listing}{tb}{lst}{}
\floatname{listing}{Listing}
\makeatletter
\gdef\affiliations_{}  
\renewcommand{\affiliations}[1]{\gdef\affiliations_{#1}}
\makeatother

\title{Neural Breadcrumbs: Membership Inference Attacks on LLMs Through Hidden State and Attention Pattern Analysis}

\author {
    Disha Makhija \textsuperscript{\rm 1},
    Manoj Ghuhan Arivazhagan \textsuperscript{\rm 1},
    Vinayshekhar Bannihatti Kumar \textsuperscript{\rm 1},
    Rashmi Gangadharaiah \textsuperscript{\rm 1}
}
\affiliations {
    \textsuperscript{\rm 1}AWS AI Labs\\
}

\begin{document}

\maketitle

\begin{abstract}

Membership inference attacks (MIAs) reveal whether specific data was used to train machine learning models, serving as important tools for privacy auditing and compliance assessment. Recent studies have reported that MIAs perform only marginally better than random guessing against large language models, suggesting that modern pre-training approaches with massive datasets may be free from privacy leakage risks. Our work offers a complementary perspective to these findings by exploring how examining LLMs' internal representations, rather than just their outputs, may provide additional insights into potential membership inference signals. Our framework, \emph{memTrace}, follows what we call \enquote{neural breadcrumbs} extracting informative signals from transformer hidden states and attention patterns as they process candidate sequences. By analyzing layer-wise representation dynamics, attention distribution characteristics, and cross-layer transition patterns, we detect potential memorization fingerprints that traditional loss-based approaches may not capture. This approach yields strong membership detection across several model families achieving average AUC scores of 0.85 on popular MIA benchmarks. Our findings suggest that internal model behaviors can reveal aspects of training data exposure even when output-based signals appear protected, highlighting the need for further research into membership privacy and the development of more robust privacy-preserving training techniques for large language models.


\end{abstract}

\section{Introduction}
Membership inference attacks (MIAs) occupy a paradoxical position in machine learning privacy. On one hand, they serve as crucial auditing tools that can determine whether copyright-protected or sensitive data was improperly used to train models offering accountability in an era of massive, often opaque data collection practices and usage. On the other hand, they pose a significant privacy threat when weaponized by adversaries seeking to uncover whether specific individuals' data was included in training sets, potentially revealing sensitive associations or characteristics about those individuals. This dual nature becomes especially concerning for large language models due to their ubiquitous deployment across critical domains and unprecedented scale that even allows them to act on behalf of the user. Advancing MIA techniques, thus, is crucial for both these reasons. High performing MIA solutions can not only help audit improper usage of data for training, but can also help in security research by precisely identifying vulnerabilities that drive the development of more robust defenses. 

Despite their importance, recent research has raised significant questions about whether effective membership inference is even possible for large language models. A growing body of evidence suggests surprising limitations in current approaches.~\citet{duan2024membership} found that across various model sizes and domains, membership inference attempts barely outperform random guessing.~\citet{das2025blind} demonstrated that MIAs attempting to identify pre-training data in LLMs perform only marginally better than blind baselines, while~\citet{maini2024llm} argued that many purportedly successful attacks may actually be detecting temporal distribution shifts rather than true data membership. This apparent resistance to membership inference has been attributed to several factors unique to modern LLMs such as their massive training datasets combined with relatively few training iterations, and the inherently blurred boundaries between member (texts that were seen by the model during training) and non-member data (texts that were not seen by the model during training) points due to substantial n-gram overlap in natural language corpora. In contrast, other work has shown that these models may be significantly more vulnerable to membership inference attacks after fine-tuning on smaller, specialized datasets~\citep{zhang2025finetuning, mia_tuner}. 

\begin{figure}
            \includegraphics[width=\columnwidth]{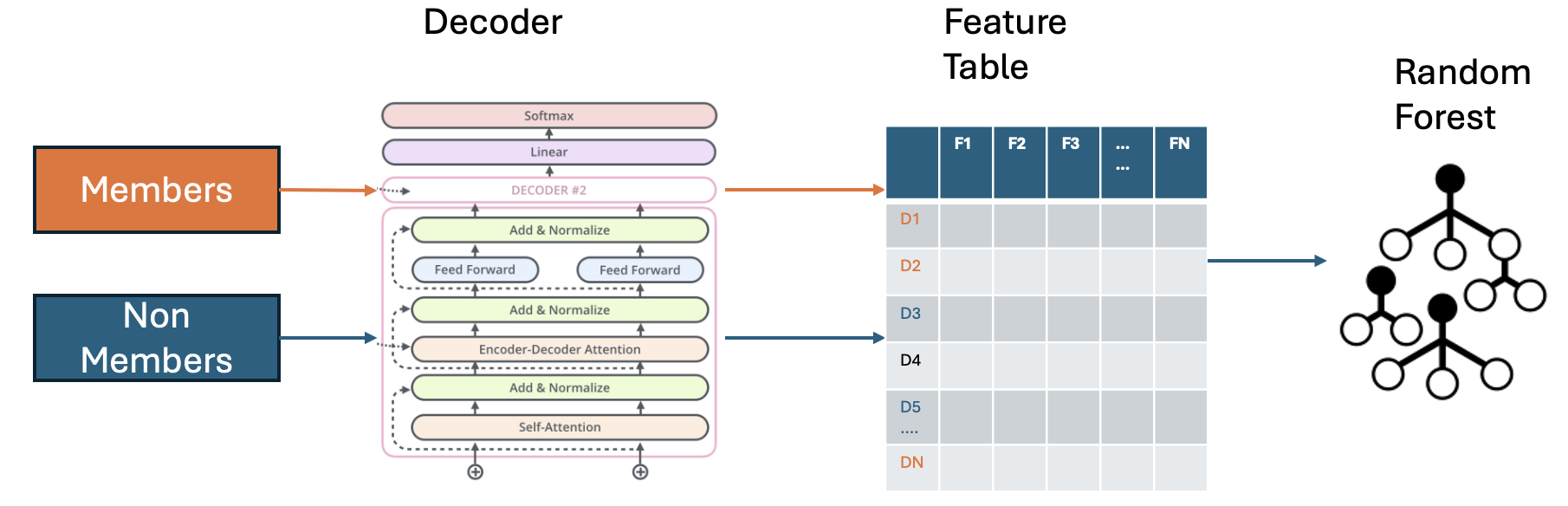}
		\caption{We leverage decoder based LLMs to generate several features for members and non members. These features are then used to train a Random Forest classifier. This classifier acts as an interpretable system to identify why a sample at inference is called as a member or a non member.  }
		\label{fig:model_arch}
\end{figure}

We propose an alternative perspective : the limitations of existing MIA approaches may stem not from the absence of membership signals but from their primary focus on model outputs which examine only the final layer logits or next-token probabilities. This conventional approach effectively reduces the entire computational process of a massive language model to a single terminal point, potentially overlooking rich information encoded within the model's internal dynamics. The processing of a sequence through many transformer layers leaves what we call \enquote{neural breadcrumbs} which are traces of how the model processes information differently when encountering member versus non-member sequences. These processing patterns, while potentially obscured in the final output distribution, may be detectable in the model's intermediate computations and attention mechanisms.

Our approach, \emph{memTrace}, leverages this insight by extracting features from hidden state representations and attention patterns across the full transformer architecture. Unlike previous methods, \emph{memTrace} does not require fine-tuning the LLM on additional (often specialized) training data, instead directly analyzing how representations evolve through the network when processing candidate sequences. By examining layer-wise representation dynamics, attention distribution characteristics, and cross-layer transition patterns, we can train lightweight classifiers like single-layer MLPs, random forests, etc. to distinguish between member and non-member sequences. This approach is summarized in Figure~\ref{fig:model_arch}. Our comprehensive evaluation across multiple model architectures and diverse text domains reveals that this approach achieves substantially stronger membership detection capabilities than output-based methods, with average AUC scores of 0.85 on popular MIA benchmarks. These findings suggest that while LLMs may appear resistant to traditional membership inference techniques, their internal processing mechanisms can still reveal subtle but detectable signs of training data exposure. 



Our \textbf{key contributions} could be summarized as follows:

\begin{enumerate}
  \item We introduce \emph{memTrace}, a membership inference approach that extracts multi-faceted features from transformer hidden states and attention patterns. Unlike previous methods, our approach analyzes model's internal representation dynamics to reveal membership signals without heavy-weight fine-tuning of the LLM.
  \item We present empirical evidence contradicting the prevailing hypothesis that modern LLM training paradigms inherently limit membership inference risk, demonstrating successful attacks across multiple model architectures and diverse text domains.
  \item We provide interpretable insights into how LLMs process familiar content differently from unfamiliar data. Our analysis reveals that membership often manifests through distinctive processing pathways for members versus non-members. These findings contribute to a mechanistic understanding of how memorization affects model processing across the transformer stack.
  \end{enumerate}

Our findings fundamentally re-frame the privacy risk landscape for large language models by identifying that analyzing the internal representations of large language models may provide additional insights into their memorization behaviors. As LLMs continue their deployment across sensitive domains from healthcare to legal applications, our work indicates that privacy risks warrant continued research attention, particularly in developing privacy-preserving training techniques.



\section{Background and Related Work}
This section surveys prior work on membership inference attacks against machine learning models, with a focus on methods applicable to large language models. We begin by defining key concepts, then organize existing approaches into reference-free and reference-based families, discuss methods tailored to LLMs and conclude by contrasting these with our proposed hidden‐state contrast framework.

A \emph{membership inference attack} (MIA) attempts to decide, for a target model $f$ and a candidate example $x$, whether $x$ was included in the model’s training set $\mathcal{D}_{\mathrm{train}}$. The adversary outputs a membership guess $\hat b \in \{0,1\}$, aiming to maximize the \emph{attack advantage}:
$$\mathrm{Adv} = \bigl|\Pr[\hat b = 1 \mid x \in \mathcal{D}_{\mathrm{train}}] - \Pr[\hat b = 1 \mid x \not\in \mathcal{D}_{\mathrm{train}}]\bigr|.$$

Membership inference attacks vary along three critical dimensions: (i) \textbf{model access} (black-box vs.\ white-box), where black-box refers to accessing only model outputs while white-box permits inspection of model's internal parameters; (ii) \textbf{reference requirements}, distinguishing between methods that require auxiliary models or datasets and those that operate independently; and (iii) \textbf{inference signals} used to detect membership, ranging from simple loss values to complex representation patterns.

\paragraph{Reference-Free Methods} The simplest membership inference approaches rely solely on the target model's behavior without external calibration. We review the key attacks in this category. Loss or Perplexity Attack, also known as the PPL or Loss attack, method identifies low loss (or perplexity) as an indicator of membership. The underlying assumption is that data points seen during training will result in lower loss values compared to unseen examples. However, this simple approach often performs poorly because non-member sequences, especially those containing frequent or repetitive text, can also achieve low perplexity~\citep{yeom2018privacyriskmachinelearning,carlini2022membershipinferenceattacksprinciples}. The Min-K\% Prob method  ~\citep{shi2024detecting} is a reference-free MIA designed to overcome the limitations of simpler attacks by focusing on the tokens within a sequence that the model is least confident about. Min–K\% PROB computes the average log‐probability of the lowest‐probability tokens, while Min–K\%++  further amplifies local maxima in the probability distribution ~\citep{zhang2024min}. These methods improve over plain PPL without requiring any reference model. The Neighborhood attack proposes using probability curvature, for each candidate $x$, generating semantically similar neighbors via perturbation and compare their average loss. Low curvature (i.e.\ small change in loss in the neighborhood) signals memorization~\citep{mattern-etal-2023-membership}. DetectGPT adapts this idea zero-shot using an external language model that infills randomly masked spans of the original text to identify non members from the training text~\citep{detectgpt}. Zlib and Lowercase attacks compare perplexity on the original $x$ to compressed or lowercased variants ~\citep{carlini2021extracting}. The intuition is that members, having been seen during training, yield smaller perplexity increases under such transformations. ~\citet{bertran2023scalable} propose a new family of MIAs that are competitive with state-of-the-art shadow model approaches like LiRA~\citep{carlini2022membershipinferenceattacksprinciples} while requiring substantially fewer computational resources and less knowledge of the target model's architecture. This work focuses on the efficiency and scalability of MIAs, identifying pinball loss as a key target objective for effective attacks.

\paragraph{Reference-Based Methods} These methods calibrate the target model’s scores against those from a reference model trained on separate data. Likelihood-Ratio Attacks (LiRA) computes a likelihood ratio between two hypotheses—“$x$ was in training” vs.\ “$x$ was not”—using scores from shadow/reference models. LiRA can achieve high precision but requires training many shadow models and a closely matched reference corpus~\citep{carlini2022membershipinferenceattacksprinciples}. Self-Prompted MIA (SPV-MIA) address LiRA’s dependence on external data by prompting the target LLM itself to generate a reference dataset. They define a probabilistic variation metric that compares score distributions on generated vs.\ real examples, improving robustness when no true reference set is available~\citep{fu2024membership}. Robust MIA (RMIA) propose a statistical test under the null hypothesis that $x$ is replaced by a random sample and then compose many pairwise likelihood ratio tests where each test compares the likelihood of the data point to the random sample. RMIA combines a small number of reference models with population samples to yield a computationally efficient and distribution-robust attack~\citep{zarifzadeh2024lowcost}. Fine-Tuned Score Deviation (FSD) fine-tune the target LLM on a small auxiliary corpus and measure how much perplexity of non-members decreases relative to members. This enlarges the membership signal and yields significant AUC gains in pre-training data detection~\citep{zhang2025finetuning}. Test-Set Contamination Detection leverage dataset exchangeability suggesting if an LLM has seen an evaluation example, it prefers the canonical ordering of lines over shuffled inputs. This black-box test serves as a “provable” contamination check for benchmarks~\citep{oren2024proving}. PII Leakage Attacks systematically probe for personally identifiable information (PII) extraction, reconstruction, and inference. They find that differential privacy (DP) and PII scrubbing reduce but do not eliminate leakage~\citep{lukas2023analyzing}. MIA Tuner introduces a novel instruction-based MIA method that enables LLMs themselves to serve as more precise pre-training data detectors by using instruction-tuning (SFT) to adapt LLMs to directly answer whether a text belongs to their pre-training dataset or not~\citep{mia_tuner}.

\paragraph{Efficacy of MIAs} Recent studies~\citep{das2025blind,maini2024llm,duan2024membership} suggest MIAs often perform barely better than random guessing on LLM pre-training data. Skeptics attribute this to massive training datasets with few iterations, blurry boundaries between member and non-member texts due to n-gram overlap in natural language corpora, and potential confounding factors like detecting temporal distribution shifts rather than true membership.~\citet{puerto2024scalingmembershipinferenceattacks} also challenge the notion that MIAs do not work on LLMs, arguing that they do work but only when multiple documents are presented for testing, rather than short sequences. They construct new benchmarks that measure MIA performance at continuous scales, from sentences to collections of documents.

In contrast, our work demonstrates that membership signals exist but are primarily encoded in internal model dynamics rather than outputs. This approach reveals detectable membership information in LLMs' processing pathways without requiring the model fine-tuning. Our results show that deeper transformer layers harbor increasingly strong membership signals, advancing both theoretical understanding and practical capabilities of membership inference in large language models.

\section{Methodology}
We introduce \emph{memTrace}, a membership inference framework that analyzes transformer hidden states and attention patterns to identify distinctive processing signatures associated with training data exposure. Unlike output-focused approaches, our method captures the subtle \enquote{neural breadcrumbs} left throughout the model's internal layers as it processes familiar versus unfamiliar content.

\paragraph{Feature Vector Construction} We create a feature vector that captures behavioral signatures across the model's hidden layers. For each input sequence $x_m$ and a transformer model $M$, we construct a feature vector $\mathbf{f}_m$ that characterizes the model's processing behavior. For each sequence, we collect hidden states $H^{(l)} \in \mathbb{R}^{n \times d}$ from each layer $l$, extract attention weights $A^{(l)} \in \mathbb{R}^{ h \times n \times n}$ from each layer $l$, and obtain logits $L^{(l)} \in \mathbb{R}^{ n \times |V|}$ from prediction layers for a sequence of size $n$ with hidden layer size as $d$ and vocabulary $V$. We then compute layer-specific features from these internal representations, analyzing how they transform and evolve. Finally, these features are aggregated into a single fixed-length vector characterizing the model's processing behavior. This pipeline generates a comprehensive feature vector that serves as a fingerprint of how the model processes each input sequence. For a model with $L$ layers, the feature vector includes statistics computed across all layers, capturing both layer-specific behaviors and cross-layer patterns.

\noindent \textbf{Layer Transition Features} The progression of representations through a transformer's layers reveals how information is incrementally transformed and refined. We quantify these dynamics through representation surprise metrics that measure the distance between hidden states of consecutive layers for each token position. For each token position $t$ and layer transition from $i$ to $i+1$, we compute the transition surprise as $\text{transition\_surprise}_{i \rightarrow i+1}[t] = \left\| \mathbf{h}^{(i+1)}_t - \mathbf{h}^{(i)}_t \right\|_2$, which measures the Euclidean distance between consecutive layer representations. We also calculate normalized version of surprise to capture directional changes independent of magnitude. Additionally, we measure representation stability using cosine similarity, $\text{stability}_{i \rightarrow i+1}[t] = \cos(\mathbf{h}^{(i+1)}_t, \mathbf{h}^{(i)}_t)$. For each layer transition, we compute statistical aggregates including mean, minimum, maximum, standard deviation, argmin, and argmax across token positions. These statistics characterize how information transforms between successive layers, potentially revealing different processing pathways for memorized versus novel content.


\noindent \textbf{Prediction Confidence and Entropy Features} For each layer that produces logits, we analyze the model's predictive behavior through entropy and confidence metrics. The entropy at each token position is calculated as $\text{entropy}[t] = -\sum_{v \in V} p_t(v) \log p_t(v)$, where $p_t(v)$ represents the softmax probability for vocabulary item $v$. Prediction confidence is measured as $\text{confidence}[t] = \max_{v \in V} p_t(v)$, representing the model's certainty in its top prediction. We also compute the confidence gap as $\text{confidence\_gap}[t] = p_t(v_1) - p_t(v_2)$, where $v_1$ and $v_2$ are the top two predicted tokens. This gap indicates how decisively the model favors its primary prediction over alternatives. For each layer, we compute statistical aggregates of these metrics across token positions, yielding features that characterize the model's certainty profile. We further analyze confidence stability as $\text{confidence\_stability} = \frac{\mu(\text{confidence})}{\sigma(\text{confidence})}$, which captures how consistently confident the model is across token positions. Higher values indicate more uniform confidence, while lower values suggest the model has varying degrees of certainty when processing different parts of the sequence, a pattern we hypothesize may correlate with partial memorization.


\noindent \textbf{Attention Pattern Analysis} For each layer $l$ and attention head $h$, we extract attention weights $A^{l,h} \in \mathbb{R}^{n \times n}$ for sequence length $n$ and analyze them across multiple dimensions. We first examine general attention distribution features by averaging across heads to obtain $\bar{A}^l$ and calculating metrics such as attention entropy: $\text{attention\_entropy} = \frac{1}{n} \sum_{t=1}^{n} \left( -\sum_{s=1}^{n} \bar{a}^l_{t,s} \log_2 (\bar{a}^l_{t,s} + \epsilon) \right)$. We also compute attention concentration as $\text{attention\_concentration} = \frac{1}{n} \sum_{t=1}^{n} \max_{s} \bar{a}^l_{t,s}$, which measures how decisively the model focuses on specific tokens. Additionally, we quantify attention sparsity as $\text{attention\_sparsity} = \text{mean}(\bar{A}^l < \tau)$, where $\tau$ is adaptively determined based on the layer's attention statistics. Lower entropy indicates more focused attention, while higher sparsity suggests more selective information routing. For head-specific analysis, we calculate metrics for each individual attention head, including head entropy $\text{head\_entropy}[h]$ and head focus $\text{head\_focus}[h] = \frac{1}{n} \sum_{t=1}^{n} \max_{s} a^{l,h}_{t,s}$. 

Position-based attention features analyze the spatial patterns in attention allocation. We measure self-attention quantifying how much tokens attend to themselves. Previous token bias is calculated as $\text{prev\_token\_bias} = \frac{1}{n-1} \sum_{t=2}^{n} \bar{a}^l_{t,t-1}$, capturing the model's tendency to focus on immediately preceding context. We also compute the mean attention distance as $\text{mean\_attention\_distance} = \frac{\sum_{t,s} |t-s| \cdot \bar{a}^l_{t,s}}{\sum_{t,s} \bar{a}^l_{t,s}}$, which characterizes whether the model tends to attend locally or globally. These features characterize the model's attention to different positional relationships, revealing how it integrates information across sequence positions. Collectively, these attention-based features provide a detailed characterization of how models allocate computational focus, revealing subtle but consistent differences between the processing of memorized versus novel content. Our implementation adaptively adjusts feature extraction based on the specific model architecture being analyzed, allowing for consistent comparison across model families.

\noindent \textbf{Context Evolution Features} We analyze how context representations evolve as more tokens are processed by computing $\text{context\_evolution}[i] = \left\| \frac{1}{i+1}\sum_{t=0}^{i+1} \mathbf{h}_t - \frac{1}{i}\sum_{t=0}^{i} \mathbf{h}_t \right\|_2$. This measures how much the average representation changes when adding each new token to the context. The statistical properties of these evolution steps reveal how efficiently the model integrates new information, with memorized content potentially showing different integration patterns 
compared to novel content.

\noindent \textbf{Token-Position Specific Features} For specific token positions (e.g., first, middle, last), we compute specialized features that capture position-dependent processing. We measure first-to-last token similarity as $\text{first\_last\_similarity} = \cos(\mathbf{h}_0, \mathbf{h}_{n-1})$, capturing the relationship between the beginning and end of the sequence. For specific token positions $k$, we compute local statistics such as $\text{token\_k\_confidence\_std} = \sigma(\{\text{confidence}[t]\}_{t=k-c}^{k+c})$, which measures confidence variation in the local neighborhood of position $k$. These position-specific metrics capture how differently the model processes tokens based on their sequential position, potentially revealing position-dependent memorization patterns.

The complete feature vector combines all metrics described above across all layers, resulting in a comprehensive representation of the model's internal processing behavior. All features are normalized using z-score standardization to ensure comparable scales across different feature types. 

\noindent \paragraph{Membership Inference Classifier}
Using the extracted features, we train a random forest classifier to predict whether an input was part of the model's training data. We implement a robust cross-validation framework to ensure reliable performance. First, we employ 5-fold stratified cross-validation with an inner hyperparameter optimization loop using RandomizedSearchCV. We explore parameters including the number of estimators (100-400), maximum tree depth (3-10), minimum samples for splits and leaf nodes, and feature selection strategies. All numerical features are standardized using StandardScaler, fitted only on training data. The final classifier is trained on 80\% of the data with the most commonly selected hyperparameters across folds, and evaluated on a held-out 20\% test set. We report performance using AUC as our primary metric due to its robustness to class imbalance following previous literature reported as baselines. Feature importance analysis provides insights into which behavioral patterns most strongly indicate memorization in large language models.

\section{Experiments}
In this section we describe the experiment setup and present our key results.

\subsection{Experiment Setup}
\paragraph{Datasets} We evaluate our approach on established membership inference benchmarks spanning diverse domains. MIMIR Benchmark~\citep{duan2024membership} provides carefully curated member/non-member pairs across multiple domains, controlling for confounding factors like temporal effects and distribution shifts. We use its Wikipedia, GitHub, PubMed Central, HackerNews, and DM Mathematics subsets obtained from The Pile, which represent different text domains with varying linguistic characteristics and structural properties. The MIMIR benchmark contains multiple variants of each dataset, each created using different filtering criteria. We specifically focus on the 13-gram $0.8$ split version, which refers to the set where the 13-gram overlap between member and nonmember texts is unrestricted and could be anywhere between 0-80\%, making it significantly more difficult for models to distinguish between the two classes. WikiMIA and BookMIA~\citep{shi2024detecting} contain Wikipedia articles and book excerpts respectively, with membership labels based on the presence of content in known training corpora. These datasets provide longer text sequences that better reflect real-world use cases.

\paragraph{Models} We evaluate our approach on three families of language models that span diverse architectures, scales, and training methodologies: i) Pythia~\citep{Biderman2023pythia} : A suite of models ranging from 70M to 6.9B parameters, all trained on The Pile dataset; ii) LLaMA v1~\citep{Touvron2023llama} Meta's 7B-parameter foundational model trained on trillions of tokens from diverse publicly available corpora. These models employ different architectural choices from Pythia, including activations and rotary positional embeddings, providing a test of our method's generalization across architectural variations; iii) GPT-Neo~\citep{gptneo} EleutherAI's 1.3B and 2.7B parameter transformers trained on The Pile using a Mesh-Tensorflow implementation, offering another test of cross-architecture generalization.

\paragraph{Baselines} We compare our approach against established membership inference methods and the current state-of-the-art in membership inference approaches, including Perplexity~\citep{yeom2018privacyriskmachinelearning} that uses raw sequence likelihood, with lower perplexity suggesting training exposure; Min-K\%~\citep{shi2024detecting} that focuses on the least confident token predictions to avoid confounding effects from common patterns; Lowercase~\citep{carlini2022membershipinferenceattacksprinciples} that computes the ratio of perplexities between original and lowercased text; Neighborhood~\citep{mattern-etal-2023-membership} that examines probability curvature around examples through controlled perturbations; Zlib~\citep{carlini2022membershipinferenceattacksprinciples} that uses the ratio between perplexity and compression entropy as a membership signal, and techniques requiring fine-tuning such as MIATuner~\citep{mia_tuner} that instruction-tunes models to directly identify their training data; Fine-tuned Score Deviation~\citep{shi2024detecting} that leverages fine-tuning on auxiliary data to amplify the gap between member and non-member scores. Since Fine-tuned score deviation method can work on top of other attacks, we compute its performance for every attack (refer to appendix) but report the numbers corresponding to the perplexity attack since they were the highest. For reproducing baselines that require tuning, we use the best hyper-parameters provided in their respective papers, note that some baseline implementations did not work in some cases and those results are marked NA in the main table. For all reference based attacks we use stablelm-base-alpha-3b-v2 as the reference model as suggested by~\citet{duan2024membership}.

\paragraph{Setup} Our experiments utilized models downloaded from the EleutherAI (for Pythia models and GPT-Neo models) and huggyllama (for LLaMA) repositories on Hugging Face. Each model was evaluated using its native tokenizer from the same repository to ensure consistency with training conditions. All experiments were conducted on an instance equipped with 8 NVIDIA A100 GPUs (40GB memory each), 96 vCPUs, and 1,152 GB of RAM. We processed samples with a batch size of 32 and padding and truncation set to True for feature extraction across all model sizes, allowing us to efficiently compute the full feature vectors within GPU memory constraints. The inference pipeline was implemented using PyTorch 1.13 with CUDA 11.6, and we utilized the Hugging Face Transformers library for model loading and tokenization. For all experiments including baselines, we employed 5-fold stratified cross-validation (random seed - 420) with to ensure robust evaluation, and all reported performance metrics represent averages across these 5 folds. We conducted variance analysis across all cross-validation folds and found consistent performance with standard deviations typically below 0.03 AUC, indicating stable results.  The variance is provided in Appendix due to space constraints.

\subsection{Results}
The main results comparing AUC of our method against the baselines across different datasets and models are included in Table~\ref{tab:auc_results}. For challenging membership inference scenarios, particularly in domains such as Wikipedia, Pubmed Central, Hacker News, and Github, our method demonstrates substantial improvements in AUC compared to all baseline approaches. On the WikiMIA dataset, while the fine-tuning based baseline methods achieve strong performance due to the dataset's distinct distributional properties, our method maintains competitive performance, matching the high AUC scores of existing approaches. For the BookMIA dataset, which contains longer sequences of 512 words, our method achieves near-perfect AUC scores. The extended sequence length provides more opportunities for detecting distinctive processing patterns, making membership inference more reliable. The performance numbers of our method on the Github dataset being slightly lower compared to that of other datasets is due to the large percentage of texts in members and non-members with 13-gram overlap. This issue is discussed in detail below. Overall, the consistency across architectures and scales of our method is particularly significant as it suggests that vulnerability to membership inference is an inherent characteristic of language models, regardless of their size or architectural design. 

\begin{table*}[t]
\setlength\tabcolsep{3pt}
\begin{center}
\begin{small}
\scalebox{0.9}{
\hspace*{-1.5cm}
\begin{tabular}{l|ccccccc|ccccccc|ccccccc}
\toprule
& \multicolumn{7}{c|}{\textbf{Wikipedia}} & \multicolumn{7}{c|}{\textbf{Pubmed Central}} & \multicolumn{7}{c}{\textbf{Hacker News}} \\
\cmidrule{2-22}
\textbf{Method} & \rotatebox{90}{Pythia-70M} & \rotatebox{90}{Pythia-410M} & \rotatebox{90}{Pythia-1B} & \rotatebox{90}{Pythia-6.9B} & \rotatebox{90}{LLaMA-7B} & \rotatebox{90}{GPTNeo-1.3B} & \rotatebox{90}{GPTNeo-2.7B}  & \rotatebox{90}{Pythia-70M} & \rotatebox{90}{Pythia-410M} & \rotatebox{90}{Pythia-1B} & \rotatebox{90}{Pythia-6.9B} & \rotatebox{90}{LLaMA-7B} & \rotatebox{90}{GPTNeo-1.3B} & \rotatebox{90}{GPTNeo-2.7B} & \rotatebox{90}{Pythia-70M} & \rotatebox{90}{Pythia-410M} & \rotatebox{90}{Pythia-1B} & \rotatebox{90}{Pythia-6.9B} & \rotatebox{90}{LLaMA-7B} & \rotatebox{90}{GPTNeo-1.3B} & \rotatebox{90}{GPTNeo-2.7B} \\
\midrule
Perplexity & 0.51  & 0.50 & 0.50 & 0.52 & 0.54 & 0.51 & 0.51 & 0.50 & 0.49 & 0.50 & 0.52 & 0.51 & 0.49 & 0.50 & 0.49 & 0.49 & 0.50 & 0.52 & 0.51 & 0.51 & 0.51 \\
Min-K\% & 0.47 & 0.48 & 0.51 & 0.51 & 0.49 & 0.52 & 0.52 & 0.51 & 0.50 & 0.51 & 0.53 & 0.53 & 0.50 & 0.51 & 0.51 & 0.50 & 0.51 & 0.52 & 0.52 & 0.51 & 0.52 \\
Lowercase & 0.51 & 0.47 & 0.49 & 0.52 & 0.48 & 0.51 & 0.50 & 0.51 & 0.51 & 0.50 & 0.52 & 0.53 & 0.52 & 0.52 & 0.50 & 0.50 & 0.51 & 0.51 & 0.52 & 0.52 & 0.52 \\
Zlib & 0.52 & 0.51 & 0.52 & 0.53 & 0.54 & 0.52 & 0.52 & 0.51 & 0.50 & 0.50 & 0.52 & 0.52 & 0.51 & 0.51 & 0.50 & 0.50 & 0.50 & 0.53 & 0.53 & 0.50 & 0.51 \\
Neighborhood & 0.53 & 0.52 & 0.52 & 0.53 & 0.54 & 0.53 & 0.53 & 0.48 & 0.48 & 0.49 & 0.53 & 0.54 & 0.51 & 0.51 & 0.50 & 0.51 & 0.50 & 0.52 & 0.52 & 0.50 & 0.51 \\
MIATuner & 0.49 & 0.51 & 0.51 & 0.49 & 0.50 & 0.50 & 0.49 & 0.50 & 0.50 & 0.48 & 0.50 & 0.50 & 0.50 & 0.52 & 0.50 & 0.50 & 0.53 & 0.50 & 0.50 & 0.49 & 0.49  \\
FSD & 0.48 & 0.47 & 0.49 & 0.47 & 0.50 & 0.50 & 0.49 & 0.49 & 0.49 & 0.49 & 0.47 & 0.48 & 0.48 & 0.49 & 0.49 & 0.49 & 0.49 & 0.47 & 0.50 & 0.49 & 0.47 \\
\textbf{memTrace} (Ours) & \textbf{0.65} & \textbf{0.87} & \textbf{0.89} & \textbf{0.89} & \textbf{0.90} & \textbf{0.86} & \textbf{0.86} & \textbf{0.59} & \textbf{0.89} & \textbf{0.88} & \textbf{0.84} & \textbf{0.80} & \textbf{0.66} & \textbf{0.72} & \textbf{0.82} & \textbf{0.81} & \textbf{0.82} & \textbf{0.83} & \textbf{0.85} & \textbf{0.71} & \textbf{0.78} \\
\midrule
\midrule
& \multicolumn{7}{c|}{\textbf{WikiMIA}} & \multicolumn{7}{c|}{\textbf{BookMIA}} & \multicolumn{7}{c}{\textbf{Github}} \\
\midrule
Perplexity & 0.58 & 0.60 & 0.60 & 0.65 & 0.64 & 0.61 & 0.63 & 0.57 & 0.59 & 0.59 & 0.61 & 0.58 & 0.59 & 0.61 & 0.61 & 0.61 & 0.63 & 0.65 & 0.67 & 0.68 & 0.70 \\
Min-K\% & 0.57 & 0.59 & 0.60 & 0.67 & 0.63 & 0.62 & 0.61 & 0.59 & 0.61 & 0.62 & 0.64 & 0.62 & 0.62 & 0.64 & 0.62 & 0.62 & 0.64 & 0.69 & 0.67 & 0.66 & 0.69 \\
Lowercase & 0.54 & 0.57 & 0.57 & 0.59 & 0.58 & 0.59 & 0.60 & 0.59 & 0.59 & 0.60 & 0.65 & 0.64 & 0.60 & 0.59 & 0.59 & 0.58 & 0.60 & 0.59 & 0.58 & 0.59 & 0.60 \\
Zlib & 0.59 & 0.60 & 0.59 & 0.62 & 0.60 & 0.62 & 0.61 & 0.60 & 0.61 & 0.61 & 0.62 & 0.63 & 0.62 & 0.61 & 0.63 & 0.66 & 0.67 & 0.70 & 0.71 & 0.70 & 0.70 \\
Neighborhood & 0.53 & 0.55 & 0.54 & 0.66 & 0.64 & 0.63 & 0.65 & 0.60 & 0.62 & 0.60 & 0.68 & 0.65 & 0.65 & 0.68 & 0.64 & 0.64 & 0.65 & 0.69 & 0.68 & 0.66 & 0.67 \\
MIATuner & 0.50 & \textbf{0.96} & \textbf{0.98} & \textbf{0.95} & 0.93 & 0.89 & \textbf{0.96} & 0.35 & 0.96 & 0.98 & 0.97 & 0.98 & N/A & N/A & 0.59 & 0.66 & 0.70 & 0.72 & 0.70 & N/A & N/A  \\
FSD & 0.81 & 0.89 & 0.92 & 0.91 & 0.90 & \textbf{0.92} & 0.92 & \textbf{0.95} & \textbf{0.98} & \textbf{0.98} & \textbf{0.98} & \textbf{0.99} & \textbf{0.98} & 0.98 & 0.45 & 0.46 & 0.42 & 0.37 & 0.49 & 0.49 & 0.47  \\
\textbf{memTrace} (Ours) & \textbf{0.82} & 0.87 & 0.80 & 0.87 & \textbf{0.94} & 0.79 & 0.85 & 0.91 & 0.94 & 0.95 & 0.95 & 0.98 & \textbf{0.98} & \textbf{0.99} & \textbf{0.67} & \textbf{0.70} & \textbf{0.73} & \textbf{0.77} & \textbf{0.78} & \textbf{0.72} & \textbf{0.72} \\
\bottomrule
\end{tabular}
}
\end{small}
\end{center}   
\caption{Comparison of test AUC scores across various methods, datasets and models.}
\label{tab:auc_results}
\end{table*}

\begin{figure}
\hfill
\subfigure[Domain: Wikipedia]{\includegraphics[width=4cm]{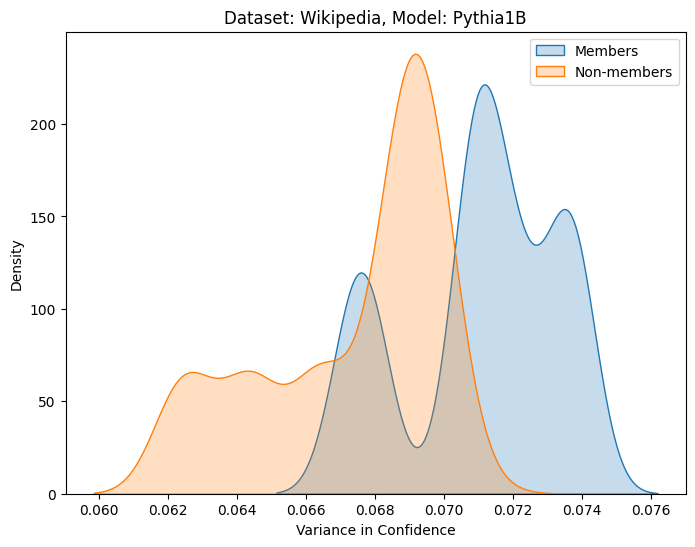}}
\hfill
\subfigure[Domain: Pubmed Central]{\includegraphics[width=4cm]{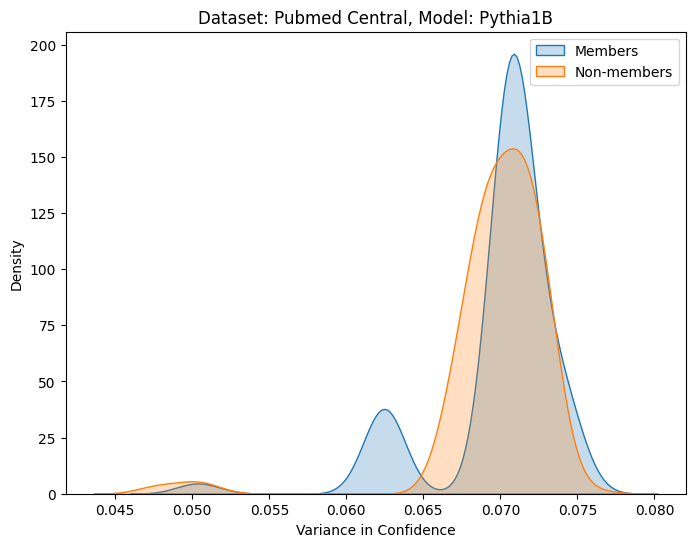}}
\hfill
\caption{Feature distribution for the variance in confidence feature for both members and non-members samples across two different data domains from the MIMIR benchmark.}
\label{fig:feat_dist}
\end{figure}

\paragraph{Feature Analysis} Our feature analysis reveals distinct processing signatures in large language models when handling previously seen versus novel texts, even though the model may not have memorized the content. We observe that when measuring the entropy of next token prediction probability distributions across each layer, previously seen content exhibits significantly higher variance in entropy compared to unseen content. This means the model's confidence fluctuates more dramatically across token positions for familiar text. This pattern suggests LLMs possibly develop recognition hot-spots, and rather than maintaining uniform confidence throughout a sequence, the model exhibits extremely high confidence at specific anchor positions where pattern recognition occurs, creating a distinctive oscillating confidence profile. Figure~\ref{fig:feat_dist} illustrates the distribution of this feature across two datasets. Additionally, we observe that certain attention heads display characteristic entropy and focus patterns when processing familiar versus unfamiliar content. The head entropy metric quantifies how uniformly attention distributes across token positions, while head focus measures how strongly an attention head concentrates on its maximum attention targets. Select heads consistently demonstrate lower entropy and higher focus for familiar content, revealing specialized mechanisms for detecting specific linguistic or statistical patterns in previously encountered text to emphasize the familiar patterns. Notably, all rules in our random forest classifier incorporate features spanning multiple network layers (early, middle, and late), rather than relying on any single layer. 

\begin{figure}
\hfill
\subfigure[Pythia1B ]{\includegraphics[width=4cm]{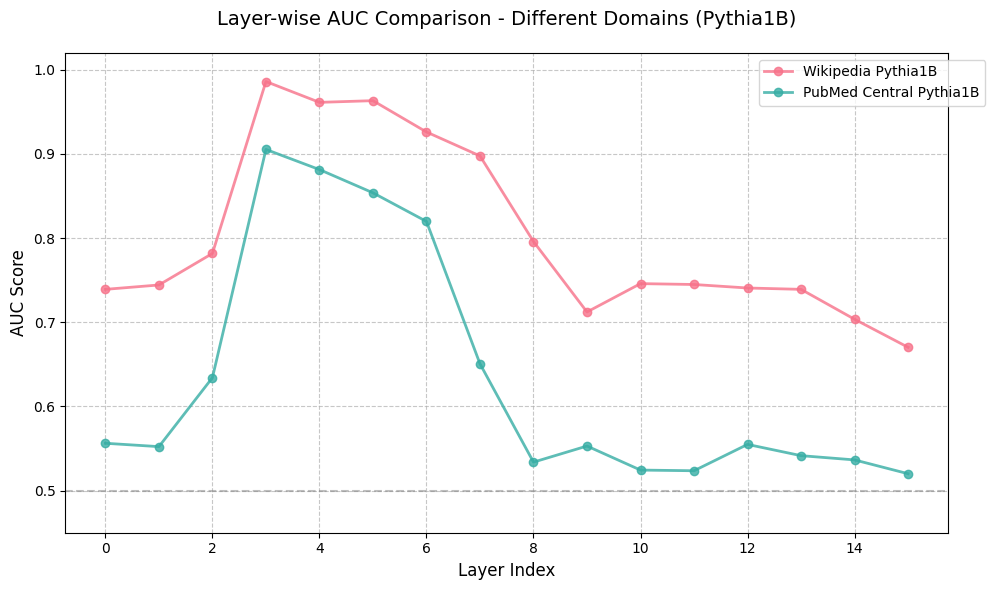}}
\hfill
\subfigure[Pythia410M]{\includegraphics[width=4cm]{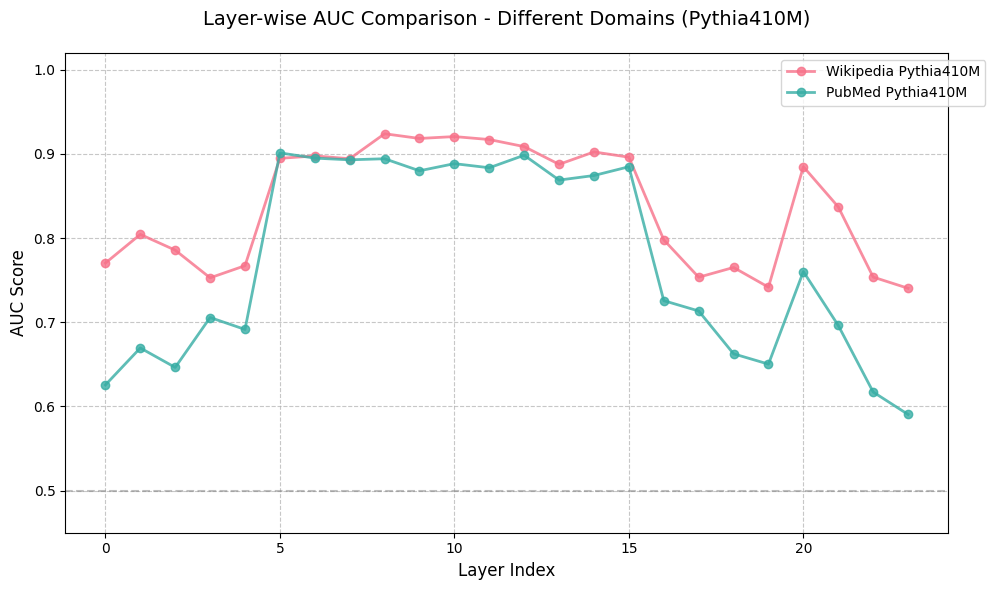}}
\hfill
\caption{Layer-wise AUC comparison on Pythia 1B and Pythia 410M models.}
\label{fig:layerwise_auc}
\end{figure}

We also compare the AUC obtained by utilizing the features from individual layers only, and show these results in Figure~\ref{fig:layerwise_auc}. For all the models and datasets, we observe that AUC is the highest at middle layers, suggesting that these layers serve as critical integration points where signatures become most detectable. It possibly suggests that in these middle layers the model has processed enough contextual information to activate specialized pathways for familiar content, but hasn't yet completed the transition to generalized output generation that occurs in later layers, indicating that memorization detection is not simply about initial feature recognition or final output prediction, but rather about identifying the distinctive processing trajectories that memorized content follows through the transformer's architecture. The lower performance at the final layer aligns with previous findings in the literature that focus on analyzing only final layer outputs and do not find differentiation between the members and the non-members.

\paragraph{MIA on Semantic Neighbors} The MIMIR benchmark provides a comprehensive dataset that includes not only members and non-members but also their semantic neighbors. These neighbors are generated through a controlled perturbation process using the BERT model with approximately 15\% masking rate. This process creates variations of the original texts while preserving their semantic meaning, for instance, replacing certain words with synonyms or restructuring sentences while maintaining the core message. The inclusion of these neighbors allows us to evaluate how membership inference attacks perform on texts that are semantically similar but not identical to the training data. We utilize our classifier originally trained to distinguish between members and non-members for testing on these neighbor samples in zero-shot way to assess whether the model's ability to detect membership extends to semantically similar content or is limited to exact matches, and report results in Table~\ref{tab:neighbors}. Based on this analysis, we observe that while the precision in identifying semantically similar members is high, the recall is still low which implies that neighbors while containing semantically similar information as members are not captured as true members by the classifier.  

\begin{table}[!htbp]
    \setlength\tabcolsep{3pt}
    \centering
    \begin{tabular}{lccc}
    \toprule
    Dataset & Model & Precision & Recall \\
    \midrule
    \multirow{2}{*}{Wikipedia} & Pythia70M & 90.1 & 26.4 \\ 
         & Pythia1B & 91 & 14.6 \\
    \multirow{2}{*}{Pubmed Central} & Pythia70M & 75.2 & 27 \\ 
         & Pythia1B & 85 & 12.9 \\
    \bottomrule
    \end{tabular}
    \caption{Membership inference performance on semantic neighbors.}
    \label{tab:neighbors}
\end{table}

\paragraph{Effect of n-gram Overlap} Different data domains naturally exhibit varying levels of n-gram overlap due to their inherent linguistic characteristics and content patterns. This variation in n-gram overlap presents a crucial challenge for membership inference: when non-member texts contain substantial subsequences that also appear in the training data (member texts), the boundary between what constitutes seen versus unseen content becomes increasingly blurred. While most domains of the Pile dataset and their samples present in the MIMIR benchmark have an 7-gram overlap of 30-40\%, two of the domains Github and DM Mathematics have much higher overlaps which exceed 73\%. Therefore, to systematically analyze the effect of this n-gram overlap on membership inference, we design a comparative experiment using two distinct subsets from these domains: a restricted subset where non-members are carefully sampled to ensure 13-gram overlap remains below 20\%, and an unrestricted subset where 13-gram overlap is allowed to reach up to 80\% (both provided in the MIMIR benchmark)~\citep{duan2024membership} and compare their performance in Table~\ref{tab:split_variation}. As expected, we observe improved membership inference performance across all model sizes when the n-gram overlap between members and non-members is more restricted. More broadly, these results suggest that domains with inherently high textual redundancy may provide natural resistance to some attacks, while domains with more unique content patterns may be more vulnerable to such attacks.

\begin{table}[t]
\centering
\begin{tabular}{l|cc|cc}
\hline
\multirow{2}{*}{Model} & \multicolumn{2}{c|}{DM Mathematics} & \multicolumn{2}{c}{Github} \\
 & < 80\% & < 20\% & < 80\% & < 20\% \\
\hline
Pythia70M & 0.63  & \textbf{0.72} & 0.67 & \textbf{0.81} \\
Pythia410M & 0.70  & \textbf{0.73} & 0..70 & \textbf{0.83} \\
Pythia1B & 0.69  & \textbf{0.72} & 0.73 & \textbf{0.84} \\
Pythia6.9B & 0.74  & \textbf{0.77} & 0.77 & \textbf{0.85} \\
\hline
\end{tabular}
\caption{Membership inference performance on high-overlap domains (Github and DM Mathematics) under different n-gram overlap constraints. Results compare restricted (<20\% 13-gram overlap) versus unrestricted (<80\% overlap) settings, demonstrating the impact of textual similarity on detection accuracy.}
\label{tab:split_variation}
\end{table}


\section{Conclusion}
In this work, we show that membership inference attacks are effective against large language models by demonstrating that membership signals are strongly encoded in models' internal representations, even when output distributions show minimal differences. Our framework achieves strong membership detection performance by analyzing how information flows through transformer architectures, revealing that models process familiar content differently than novel content. Through comprehensive evaluation across multiple model architectures and diverse text domains, we show that these internal processing signatures provide more reliable membership detection than traditional output-based approaches. These findings have important implications for privacy research in large language models: first, they demonstrate that privacy auditing must consider internal model dynamics rather than focusing solely on outputs; second, they suggest that effective privacy-preserving training techniques will need to address how models process information, not just regulate their final predictions. As language models continue to be deployed in privacy-sensitive domains, our work highlights the need for more sophisticated approaches to understanding and protecting against potential memorization vulnerabilities. A related challenge in the field is the absence of a standardized metric to evaluate the efficacy of machine un-learning techniques. This is largely due to the wide variety of datasets and incomparable benchmarks used across studies, making it difficult to assess whether a model has truly forgotten a specific piece of data. Our framework may contribute to addressing this gap. 

While this framework only works for white-box models, future work could examine whether similar processing signatures can be detected through black-box probing techniques, investigate whether these patterns generalize to other architecture families beyond decoder-only transformers, and study how fine-tuning affects these internal memorization signatures. Additionally, understanding the relationship between these processing patterns and other forms of information leakage could provide valuable directions for improving model security and privacy-preserving training methods.

\clearpage
\newpage
\bibliography{references}

\begin{thebibliography}{21}
\providecommand{\natexlab}[1]{#1}

\bibitem[{Bertran et~al.(2023)Bertran, Tang, Roth, Kearns, Morgenstern, and Wu}]{bertran2023scalable}
Bertran, M.~A.; Tang, S.; Roth, A.; Kearns, M.; Morgenstern, J.~H.; and Wu, S. 2023.
\newblock Scalable Membership Inference Attacks via Quantile Regression.
\newblock In \emph{Thirty-seventh Conference on Neural Information Processing Systems}.

\bibitem[{Biderman et~al.(2023)Biderman, Schoelkopf, Anthony, Bradley, O'Brien, Hallahan, Khan, Purohit, Prashanth, Raff, Skowron, Sutawika, and Van Der~Wal}]{Biderman2023pythia}
Biderman, S.; Schoelkopf, H.; Anthony, Q.; Bradley, H.; O'Brien, K.; Hallahan, E.; Khan, M.~A.; Purohit, S.; Prashanth, U.~S.; Raff, E.; Skowron, A.; Sutawika, L.; and Van Der~Wal, O. 2023.
\newblock Pythia: a suite for analyzing large language models across training and scaling.
\newblock In \emph{Proceedings of the 40th International Conference on Machine Learning}, ICML'23. JMLR.org.

\bibitem[{Black et~al.(2021)Black, Gao, Wang, Leahy, and Mesh-Tensorflow~Biderman}]{gptneo}
Black, S.; Gao, L.; Wang, P.; Leahy, C.; and Mesh-Tensorflow~Biderman, S. 2021.
\newblock {GPT-Neo: Large Scale Autoregressive Language Modeling with Mesh-Tensorflow}.

\bibitem[{Carlini et~al.(2022)Carlini, Chien, Nasr, Song, Terzis, and Tramer}]{carlini2022membershipinferenceattacksprinciples}
Carlini, N.; Chien, S.; Nasr, M.; Song, S.; Terzis, A.; and Tramer, F. 2022.
\newblock Membership Inference Attacks From First Principles.
\newblock arXiv:2112.03570.

\bibitem[{Carlini et~al.(2021)Carlini, Tramer, Wallace, Jagielski, Herbert-Voss, Lee, Roberts, Brown, Song, Erlingsson et~al.}]{carlini2021extracting}
Carlini, N.; Tramer, F.; Wallace, E.; Jagielski, M.; Herbert-Voss, A.; Lee, K.; Roberts, A.; Brown, T.; Song, D.; Erlingsson, U.; et~al. 2021.
\newblock Extracting training data from large language models.
\newblock In \emph{30th USENIX security symposium (USENIX Security 21)}, 2633--2650.

\bibitem[{Das, Zhang, and Tram{\`e}r(2025)}]{das2025blind}
Das, D.; Zhang, J.; and Tram{\`e}r, F. 2025.
\newblock Blind Baselines Beat Membership Inference Attacks for Foundation Models.

\bibitem[{Duan et~al.(2024)Duan, Suri, Mireshghallah, Min, Shi, Zettlemoyer, Tsvetkov, Choi, Evans, and Hajishirzi}]{duan2024membership}
Duan, M.; Suri, A.; Mireshghallah, N.; Min, S.; Shi, W.; Zettlemoyer, L.; Tsvetkov, Y.; Choi, Y.; Evans, D.; and Hajishirzi, H. 2024.
\newblock Do Membership Inference Attacks Work on Large Language Models?
\newblock In \emph{Conference on Language Modeling (COLM)}.

\bibitem[{Fu et~al.(2024)Fu, Wang, Gao, Liu, Li, and Jiang}]{fu2024membership}
Fu, W.; Wang, H.; Gao, C.; Liu, G.; Li, Y.; and Jiang, T. 2024.
\newblock Membership Inference Attacks against Fine-tuned Large Language Models via Self-prompt Calibration.
\newblock In \emph{The Thirty-eighth Annual Conference on Neural Information Processing Systems}.

\bibitem[{Fu et~al.(2025)Fu, Wang, Gao, Liu, Li, and Jiang}]{mia_tuner}
Fu, W.; Wang, H.; Gao, C.; Liu, G.; Li, Y.; and Jiang, T. 2025.
\newblock {MIA}-Tuner: Adapting Large Language Models as Pre-training Text Detector.
\newblock In \emph{Proceedings of the AAAI Conference on Artificial Intelligence}. Philadelphia, Pennsylvania, USA.

\bibitem[{Lukas et~al.(2023)Lukas, Salem, Sim, Tople, Wutschitz, and Zanella-B{\'e}guelin}]{lukas2023analyzing}
Lukas, N.; Salem, A.; Sim, R.; Tople, S.; Wutschitz, L.; and Zanella-B{\'e}guelin, S. 2023.
\newblock Analyzing Leakage of Personally Identifiable Information in Language Models.
\newblock In \emph{2023 IEEE Symposium on Security and Privacy (SP)}, 346--363. IEEE Computer Society.

\bibitem[{Maini et~al.(2024)Maini, Jia, Papernot, and Dziedzic}]{maini2024llm}
Maini, P.; Jia, H.; Papernot, N.; and Dziedzic, A. 2024.
\newblock {LLM} Dataset Inference: Did you train on my dataset?
\newblock In \emph{The Thirty-eighth Annual Conference on Neural Information Processing Systems}.

\bibitem[{Mattern et~al.(2023)Mattern, Mireshghallah, Jin, Schoelkopf, Sachan, and Berg-Kirkpatrick}]{mattern-etal-2023-membership}
Mattern, J.; Mireshghallah, F.; Jin, Z.; Schoelkopf, B.; Sachan, M.; and Berg-Kirkpatrick, T. 2023.
\newblock Membership Inference Attacks against Language Models via Neighbourhood Comparison.
\newblock In \emph{Findings of the Association for Computational Linguistics: ACL 2023}, 11330--11343. Toronto, Canada: Association for Computational Linguistics.

\bibitem[{Mitchell et~al.(2023)Mitchell, Lee, Khazatsky, Manning, and Finn}]{detectgpt}
Mitchell, E.; Lee, Y.; Khazatsky, A.; Manning, C.~D.; and Finn, C. 2023.
\newblock DetectGPT: zero-shot machine-generated text detection using probability curvature.
\newblock In \emph{Proceedings of the 40th International Conference on Machine Learning}, ICML'23. JMLR.org.

\bibitem[{Oren et~al.(2024)Oren, Meister, Chatterji, Ladhak, and Hashimoto}]{oren2024proving}
Oren, Y.; Meister, N.; Chatterji, N.~S.; Ladhak, F.; and Hashimoto, T. 2024.
\newblock Proving Test Set Contamination in Black-Box Language Models.
\newblock In \emph{The Twelfth International Conference on Learning Representations}.

\bibitem[{Puerto et~al.(2024)Puerto, Gubri, Yun, and Oh}]{puerto2024scalingmembershipinferenceattacks}
Puerto, H.; Gubri, M.; Yun, S.; and Oh, S.~J. 2024.
\newblock Scaling Up Membership Inference: When and How Attacks Succeed on Large Language Models.
\newblock arXiv:2411.00154.

\bibitem[{Shi et~al.(2024)Shi, Ajith, Xia, Huang, Liu, Blevins, Chen, and Zettlemoyer}]{shi2024detecting}
Shi, W.; Ajith, A.; Xia, M.; Huang, Y.; Liu, D.; Blevins, T.; Chen, D.; and Zettlemoyer, L. 2024.
\newblock Detecting Pretraining Data from Large Language Models.
\newblock In \emph{The Twelfth International Conference on Learning Representations}.

\bibitem[{Touvron et~al.(2023)Touvron, Lavril, Izacard, Martinet, Lachaux, Lacroix, Rozière, Goyal, Hambro, Azhar, Rodriguez, Joulin, Grave, and Lample}]{Touvron2023llama}
Touvron, H.; Lavril, T.; Izacard, G.; Martinet, X.; Lachaux, M.-A.; Lacroix, T.; Rozière, B.; Goyal, N.; Hambro, E.; Azhar, F.; Rodriguez, A.; Joulin, A.; Grave, E.; and Lample, G. 2023.
\newblock LLaMA: Open and Efficient Foundation Language Models.
\newblock arXiv:2302.13971.

\bibitem[{Yeom et~al.(2018)Yeom, Giacomelli, Fredrikson, and Jha}]{yeom2018privacyriskmachinelearning}
Yeom, S.; Giacomelli, I.; Fredrikson, M.; and Jha, S. 2018.
\newblock Privacy Risk in Machine Learning: Analyzing the Connection to Overfitting.
\newblock arXiv:1709.01604.

\bibitem[{Zarifzadeh, Liu, and Shokri(2024)}]{zarifzadeh2024lowcost}
Zarifzadeh, S.; Liu, P. C.-J.~M.; and Shokri, R. 2024.
\newblock Low-Cost High-Power Membership Inference by Boosting Relativity.

\bibitem[{Zhang et~al.(2025)Zhang, Zhang, Jing, and Wei}]{zhang2025finetuning}
Zhang, H.; Zhang, S.; Jing, B.; and Wei, H. 2025.
\newblock Fine-tuning can Help Detect Pretraining Data from Large Language Models.
\newblock In \emph{The Thirteenth International Conference on Learning Representations}.

\bibitem[{Zhang et~al.(2024)Zhang, Sun, Yeats, Ouyang, Kuo, Zhang, Yang, and Li}]{zhang2024min}
Zhang, J.; Sun, J.; Yeats, E.; Ouyang, Y.; Kuo, M.; Zhang, J.; Yang, H.~F.; and Li, H. 2024.
\newblock Min-k\%++: Improved baseline for detecting pre-training data from large language models.
\newblock \emph{arXiv preprint arXiv:2404.02936}.

\end{thebibliography}
\clearpage
\newpage
\appendix
\twocolumn[\begin{center}
\textbf{\Large Supplement for \enquote{Neural Breadcrumbs: Membership Inference Attacks on LLMs Through Hidden State and Attention Pattern Analysis}}
\end{center}]

\section{Dataset Details} \label{app:dataset_details} 
We evaluate our approach on established membership inference benchmarks spanning diverse domains. The MIMIR benchmark~\citep{duan2024membership} serves as a unified repository for evaluating membership inference attacks on LLMs. It provides a unified benchmark package that includes all existing MIAs, supporting future work. The MIMIR benchmark is constructed out of The Pile dataset, an 800GB collection of diverse text specifically curated for language modeling. Experiments within MIMIR utilize seven distinct data sources from The Pile, including knowledge sources (Wikipedia), academic papers (PubMed Central), dialogues (HackerNews), and specialized domains (DM Math, Github). For each domain, the benchmark consistently samples 1000 members from The Pile's training set and 1000 non-members from its test set. For our evaluation, we utilize specific subsets from the MIMIR benchmark that are derived from The Pile dataset. These include Wikipedia, GitHub, PubMed Central, HackerNews, and DM Mathematics. These domains were chosen because they represent diverse text sources with varying linguistic characteristics and structural properties. Members and non-members for these datasets are sampled from the train and test sets of The Pile, respectively. The MIMIR benchmark is designed to reflect the inherent difficulty of MIAs against LLMs. When referring to the 13-gram 0.8 split version, this indicates that the benchmark examples used have undergone a default decontamination process where non-member samples with greater than 80\% 13-gram overlap with the training data were filtered out. However, even with this decontamination, natural language documents commonly exhibit repeating text and substantial n-gram overlap between members and non-members. For instance, domains like Wikipedia, ArXiv, and PubMed Central, even after deduplication, show average 7-gram overlaps exceeding 30\% with the training data. This inherently fuzzy boundary and high n-gram overlap significantly decreases MIA performance for most settings across varying LLM sizes and domains. The sources suggest that non-members with lower n-gram overlap are more distinguishable by existing MIAs, and strictly resampling non-members to ensure very low n-gram overlap (e.g., $\leq$ 20\%) can lead to a significant increase in MIA performance, as discussed in Section 4 of the paper. 

\begin{table*}[!htb]
\setlength\tabcolsep{4pt}
\begin{center}
\begin{small}
\scalebox{0.9}{
\begin{tabular}{l|cccc|cccc}
\toprule
& \multicolumn{4}{c|}{\textbf{Wikipedia}} & \multicolumn{4}{c}{\textbf{Pubmed Central}} \\
\cmidrule{2-9}
\textbf{Method} & \rotatebox{90}{Pythia-70M} & \rotatebox{90}{Pythia-410M} & \rotatebox{90}{Pythia-1B} & \rotatebox{90}{Pythia-6.9B} & \rotatebox{90}{Pythia-70M} & \rotatebox{90}{Pythia-410M} & \rotatebox{90}{Pythia-1B} & \rotatebox{90}{Pythia-6.9B} \\
\midrule
MIATuner & 0.49±0.03 & 0.52±0.01 & 0.51±0.02 & 0.50±0.03 & 0.51±0.02 & 0.50±0.01 & 0.49±0.02 & 0.51±0.02 \\
FSD (PPL) & 0.49±0.03 & 0.48±0.02 & 0.50±0.03 & 0.48±0.04 & 0.49±0.03 & 0.50±0.03 & 0.49±0.03 & 0.47±0.03 \\
FSD (Lowercase) & 0.49±0.03 & 0.48±0.02 & 0.49±0.03 & 0.49±0.03 & 0.48±0.02 & 0.48±0.01 & 0.51±0.02 & 0.52±0.01 \\
FSD (Zlib) & 0.49±0.03 & 0.48±0.01 & 0.51±0.03 & 0.48±0.02 & 0.49±0.03 & 0.50±0.03 & 0.49±0.04 & 0.47±0.03 \\
FSD (Min-K\%) & 0.50±0.04 & 0.48±0.02 & 0.49±0.03 & 0.47±0.04 & 0.50±0.02 & 0.50±0.03 & 0.49±0.03 & 0.49±0.02 \\
\textbf{memTrace} (Ours) & \textbf{0.65±0.05} & \textbf{0.87±0.01} & \textbf{0.89±0.02} & \textbf{0.89±0.01} & \textbf{0.59±0.04} & \textbf{0.89±0.02} & \textbf{0.88±0.02} & \textbf{0.84±0.01} \\
\midrule
\midrule
& \multicolumn{4}{c|}{\textbf{Hacker News}} & \multicolumn{4}{c}{\textbf{Github}} \\
\midrule
MIATuner & 0.51±0.02 & 0.51±0.01 & 0.53±0.02 & 0.50±0.04 & 0.60±0.02 & 0.67±0.02 & 0.71±0.03 & 0.73±0.02 \\
FSD (PPL) & 0.50±0.01 & 0.50±0.03 & 0.50±0.02 & 0.48±0.04 & 0.46±0.02 & 0.47±0.01 & 0.42±0.03 & 0.37±0.03 \\
FSD (Lowercase) & 0.49±0.03 & 0.49±0.03 & 0.50±0.02 & 0.50±0.02 & 0.50±0.02 & 0.52±0.03 & 0.46±0.02 & 0.48±0.03 \\
FSD (Zlib) & 0.50±0.02 & 0.50±0.04 & 0.50±0.02 & 0.47±0.03 & 0.49±0.02 & 0.49±0.02 & 0.45±0.04 & 0.42±0.04 \\
FSD (Min-K\%) & 0.51±0.02 & 0.50±0.04 & 0.49±0.02 & 0.48±0.04 & 0.49±0.02 & 0.47±0.03 & 0.42±0.04 & 0.40±0.03 \\
\textbf{memTrace} (Ours) & \textbf{0.82±0.04} & \textbf{0.81±0.03} & \textbf{0.82±0.04} & \textbf{0.83±0.01} & \textbf{0.67±0.03} & \textbf{0.70±0.03} & \textbf{0.73±0.03} & \textbf{0.77±0.01} \\
\bottomrule
\end{tabular}
}
\end{small}
\end{center}   
\caption{Comparison of test AUC scores (mean ± standard deviation) across methods and datasets using Pythia models of varying sizes. FSD variants use different base methods: perplexity (PPL), lowercase perplexity, Zlib compression ratio, and Min-K\% probability. Higher values indicate better membership inference performance.}
\label{tab:additional_results}
\end{table*}

\section{Additional Results}
This section presents comprehensive experimental results that could not be included in the main paper due to space constraints. We first outline our data pre-processing methodology to ensure reproducibility. Then, we present additional experimental results that expand our analysis in three key directions: First, we provide a detailed comparison of our method against all competitive baselines across the challenging benchmark datasets (Wikipedia, PubMed Central, Hacker News, GitHub) and across all Pythia model variants (from Pythia-70M to Pythia-6.9B). These extended results include statistical significance metrics for all comparisons, which could not be accommodated in the main paper's tables. Second, we broaden our evaluation to include two additional datasets with distinct characteristics: DM Mathematics from MIMIR benchmark, which has high n-gram overlap between members and non-members (greater than 70\%) and features highly structured formal content with specialized notation, and arXivTection, which contains technical scientific writing. These datasets allow us to assess how membership inference attacks perform across diverse specialized domains with different linguistic patterns and vocabulary distributions. Third, we expand on the fine-tuned score deviation (FSD) baseline analysis. While the main paper presented results for this approach combined only with the perplexity attack (the most competitive configuration). This includes combinations with Zlib compression ratio, lowercase perplexity, and Min-K\% scoring functions, offering insights into which underlying mechanisms benefit most from the fine-tuning approach across our expanded set of datasets. The results for existing datasets are included in Table~\ref{tab:additional_results} and the new datasets are included in Table~\ref{tab:additional_results2}.

\textbf{Data Pre-processing} For data-preprocessing for all experiments, we processed input sequences using each model's native tokenizer with standardized parameters. Sequences were truncated to a maximum length of 512 tokens across all datasets, with the exception of BookMIA where we extended this limit to 1024 tokens to accommodate its longer text passages. We applied consistent padding to achieve uniform sequence lengths within batches (batch size = 32). When extracting features for our membership inference method, we carefully respected the attention masks generated during preprocessing, ensuring that padded tokens did not contribute to our feature calculations.  All preprocessing configurations were carefully matched to those used during the models' pre-training phases to ensure evaluation validity. 

Together, these supplementary results provide a stronger empirical support for the conclusions drawn in the main paper.

\begin{table*}[t]
\setlength\tabcolsep{4pt}
\begin{center}
\begin{small}
\scalebox{0.9}{
\begin{tabular}{l|cccc|cccc}
\toprule
& \multicolumn{4}{c|}{\textbf{DM Mathematics}} & \multicolumn{4}{c}{\textbf{arXivTection}} \\
\cmidrule{2-9}
\textbf{Method} & \rotatebox{90}{Pythia-70M} & \rotatebox{90}{Pythia-410M} & \rotatebox{90}{Pythia-1B} & \rotatebox{90}{Pythia-6.9B} & \rotatebox{90}{Pythia-70M} & \rotatebox{90}{Pythia-410M} & \rotatebox{90}{Pythia-1B} & \rotatebox{90}{Pythia-6.9B} \\
\midrule
MIATuner & 0.49±0.04 & 0.50±0.03 & 0.50±0.02 & 0.49±0.02 & 0.60±0.02 & 0.89±0.01 & \textbf{0.99±0.01} & 0.96±0.01 \\
FSD (PPL) & 0.51±0.01 & 0.50±0.04 & 0.48±0.03 & 0.49±0.03 & 0.77±0.05 & \textbf{0.94±0.01} & 0.96±0.00 & \textbf{0.98±0.00} \\
FSD (Lowercase) & 0.50±0.02 & 0.51±0.03 & 0.49±0.02 & 0.50±0.04 & 0.63±0.04 & 0.81±0.02 & 0.84±0.01 & 0.86±0.01 \\
FSD (Zlib) & 0.51±0.01 & 0.50±0.03 & 0.47±0.03 & 0.49±0.02 & 0.76±0.04 & \textbf{0.94±0.01} & 0.96±0.00 & \textbf{0.98±0.00} \\
FSD (Min-K\%) & 0.49±0.02 & 0.51±0.02 & 0.49±0.03 & 0.50±0.02 & 0.71±0.05 & 0.87±0.01 & 0.91±0.00 & 0.93±0.01 \\
\textbf{memTrace} (Ours) & \textbf{0.62±0.03} & \textbf{0.69±0.03}  & \textbf{0.69±0.03} & \textbf{0.74±0.02} & \textbf{0.80±0.02} & 0.91±0.01 & 0.92±0.01 & 0.94±0.02 \\
\bottomrule
\end{tabular}
}
\end{small}
\end{center}   
\caption{Additional datasets evaluation: Comparison of test AUC scores (mean ± standard deviation) on DM Mathematics and arXivTection datasets.}
\label{tab:additional_results2}
\end{table*}

\end{document}